\documentclass[sigconf]{acmart}
\usepackage{hyperref}  
\usepackage{hyperxmp}

\usepackage{appendix}
\usepackage{multirow}
\usepackage{caption}   
\usepackage{graphicx}  
\usepackage{caption}
\usepackage{subfigure} 
\usepackage{enumitem}
\usepackage{color,colortbl}

\usepackage{amsmath}

\usepackage{mathtools}
\usepackage{amsthm}

\usepackage{booktabs}
\usepackage{algorithm}
\usepackage{algpseudocode}
\usepackage{caption}
\usepackage{graphicx}
\usepackage{tikz}
\usepackage{multirow}
\usepackage{adjustbox}
\usepackage{amsmath}
\usepackage{amsfonts}
\usepackage{bm}
\definecolor{Lightgray}{rgb}{0.88,1,1}

\newcommand{\diffusionInd}{{n}}
\newcommand{\diffusionLast}{{N}}

\def\eqref#1{equation~\ref{#1}}

\def\1{\bm{1}}

\def\vzero{{\bm{0}}}

\def\vmu{{\bm{\mu}}}

\def\vs{{\bm{s}}}

\def\vx{{\bm{x}}}

\def\vz{{\bm{z}}}

\def\mI{{\bm{I}}}

\DeclareMathAlphabet{\mathsfit}{\encodingdefault}{\sfdefault}{m}{sl}
\SetMathAlphabet{\mathsfit}{bold}{\encodingdefault}{\sfdefault}{bx}{n}

\DeclareMathOperator*{\argmax}{arg\,max}

\everymath{\displaystyle}
\theoremstyle{plain}

\theoremstyle{definition}

\theoremstyle{remark}

\AtBeginDocument{%
  \providecommand\BibTeX{{%
    \normalfont B\kern-0.5em{\scshape i\kern-0.25em b}\kern-0.8em\TeX}}}


\copyrightyear{2024}
\acmYear{2024}
\setcopyright{acmlicensed}
\acmConference[KDD '24] {Proceedings of the 30th ACM SIGKDD Conference on Knowledge Discovery and Data Mining }{August 25--29, 2024}{Barcelona, Spain.}
\acmBooktitle{Proceedings of the 30th ACM SIGKDD Conference on Knowledge Discovery and Data Mining (KDD '24), August 25--29, 2024, Barcelona, Spain}
\acmISBN{979-8-4007-0490-1/24/08}
\acmDOI{10.1145/3637528.3671863}
\begin{document}

\title{Latent Conditional Diffusion-based Data Augmentation for Continuous-Time Dynamic Graph Model}

\author{Yuxing Tian}
\email{tianyxx@gmail.com}
\orcid{0009-0008-0313-5089}
\affiliation{%
  \institution{International Digital Economy Academy, IDEA Research}
  \city{Shenzhen}
  \country{China}
}

\author{Yiyan Qi}
\email{qiyiyan@idea.edu.cn}
\affiliation{%
  \institution{International Digital Economy Academy, IDEA Research}
  \city{Shenzhen}
  \country{China}
}

\author{Aiwen Jiang}
\email{jiangaiwen@jxnu.edu.cn}
\affiliation{%
  \institution{Jiangxi Normal University}
  \city{Nanchang}
  \country{China}
}

\author{Qi Huang}
\email{huangqi@jxnu.edu.cn}
\affiliation{%
  \institution{Jiangxi Normal University}
  \city{Nanchang}
  \country{China}
}

\author{Jian Guo$^\ddag$}
\email{guojian@idea.edu.cn}
\affiliation{%
  \institution{International Digital Economy Academy, IDEA Research}
  \city{Shenzhen}
  \country{China}
}

\thanks{$^\ddag$ Corresponding Author.}

\renewcommand{\shortauthors}{Yuxing Tian, Aiwen Jiang, Qi Huang, Jian Guo, \& Yiyan Qi}

\begin{abstract}
Continuous-Time Dynamic Graph (CTDG) precisely models evolving real-world relationships, drawing heightened interest in dynamic graph learning across academia and industry. However, existing CTDG models encounter challenges stemming from noise and limited historical data. Graph Data Augmentation (GDA) emerges as a critical solution, yet current approaches primarily focus on static graphs and struggle to effectively address the dynamics inherent in CTDGs. Moreover, these methods often demand substantial domain expertise for parameter tuning and lack theoretical guarantees for augmentation efficacy. To address these issues, we propose Conda, a novel latent diffusion-based GDA method tailored for CTDGs. Conda features a sandwich-like architecture, incorporating a Variational Auto-Encoder (VAE) and a conditional diffusion model, aimed at generating enhanced historical neighbor embeddings for target nodes. Unlike conventional diffusion models trained on entire graphs via pre-training, Conda requires historical neighbor sequence embeddings of target nodes for training, thus facilitating more targeted augmentation. We integrate Conda into the CTDG model and adopt an alternating training strategy to optimize performance. Extensive experimentation across six widely used real-world datasets showcases the consistent performance improvement of our approach, particularly in scenarios with limited historical data.
\end{abstract}




\maketitle

\section{Introduction}
Continuous-Time Dynamic Graphs (CTDGs), with every edge (event) having a timestamp to denote its occurrence time, are prevalent in real-world applications such as social networks~\cite{ADSN,TECDSN}, physical systems~\cite{kazemi2020representation} and e-commerce~\cite{DynTrans}.
Recently, CTDG models~\cite{GSNOP,liang2022autoregressive,CAWN,NTW,chang2020continuous,huang2020learning,JODIE,DyRep,decoupled} have gained increasing attention due to their significant representation capacity by directly learning the representation of the continuously occurring events in CTDGs. Despite the rapid advancements in CTDG models, they encounter two primary challenges. Firstly, the "observed" CTDG often falls short of accurately representing the true underlying process it intends to model, mainly due to various factors such as measurement inaccuracies, thresholding errors, or human mistakes~\cite{kumar2018false}. 
Secondly, most CTDG methods typically rely on extensive historical data for effective training ~\cite{GSNOP}. However, in many applications, obtaining such data is impractical, particularly in scenarios with a cold start. For instance, a nascent trading platform may only possess a few days' worth of user-asset interactions, rendering existing CTDG models trained on such limited data inadequate and resulting in sub-optimal performance.


Graph Data Augmentation (GDA) has emerged as a promising solution, with existing methods falling into two main categories~\cite{GDA}: structure-oriented and feature-oriented methods. Structure-oriented methods, such as ~\cite{rong2020dropedge,JOAO,graphmae}, typically involve adjusting graph connectivity by adding or removing edges or nodes. On the other hand, feature-oriented methods, exemplified by works from ~\cite{verma2021graphmix,guo2021ifmixup,han2022g,graphmae, kong2020flag}, directly modify or create raw features of nodes or edges in graphs.


However, most existing GDA methods focus on static graphs and it is challenging to directly apply on CTDG.
(1) Existing structure-oriented GDA methods heavily rely on domain knowledge, necessitating the selection of diverse augmentation strategy combinations tailored to specific graph datasets, as highlighted in~\cite{Label-invariant}.
Furthermore, structure-oriented GDA methods present inherent challenges in calibrating the extent of data augmentation, potentially leading to either over-augmentation or under-augmentation~\cite{consistency}.
This issue arises from input data processed by mapping-agnostic deep neural networks, which generate features without tailored calibration.
(2) Existing feature-oriented GDA methods concentrate on transforming node/edge features and have demonstrated significant optimization improvements with appropriate feature augmentations. However, they cannot handle graphs lacking node features or even edge features.
(3) Although Wang et al.~\cite{wang2021adaptive} propose MeTA for CTDG, this method is tailored for CTDG models~\cite{TGN,DyRep} with memory modules, limiting its compatibility with other state-of-the-art models such as DyGFormer~\cite{DyGFormer} and GraphMixer~\cite{graphmixer}.

To address these above challenges, we introduce a novel latent \textbf{\underline{cond}}itional \textbf{\underline{d}}iffusion-based \textbf{\underline{d}}ata \textbf{\underline{a}}ugmentation method for CTDG models, named \textbf{\textit{Conda}}. 
Conda adopts a sandwich-like architecture, comprising a Variational Auto-Encoder (VAE) and a conditional diffusion model. 
This design aims to create new historical neighbor embeddings derived from existing neighbor sequences, enhancing subsequent training of the CTDG model. 
Unlike traditional diffusion models trained via pre-training on the entire graphs, Conda necessitates historical neighbor sequence embeddings of target nodes for training. 
Consequently, we integrate Conda into the CTDG model and implement an alternating training strategy.
Our contributions can be summarized as follows:
\begin{itemize}[leftmargin=*]
\item We present Conda, an innovative data augmentation technique aimed at enhancing the CTDG model. This method leverages a latent conditional diffusion model to generate historical neighbor embeddings for the target node during the training phase.
\item Rather than directly manipulating the raw graph structure, Conda operates within the latent space, where it is more likely to encounter authentic samples.
\item We extensively evaluate our method on six widely used real-world datasets and compare it against seven baselines. Our results demonstrate that Conda enhances the performance of link prediction tasks on these baselines by up to 5\%. Notably, this improvement is achieved without the need for domain-specific knowledge.

\end{itemize}


\section{Preliminaries}

\subsection{Continuous-Time Dynamic Graph}
A CTDG $G$ can be represented as a chronological sequence of interactions between specific node pairs: $G=\{(u_0,v_0,t_0),\ldots, (u_n,v_n,t_n)\}$, where $t_i$ denotes the timestamp and the timestamps are ordered as $(0\le t_0\le t_1\le ...\le t_n)$. $u_i, v_i \in V$ denote the node IDs of the $i-th$ interaction at timestamp $t_i$, $V$ is the entire node set. Each node $v \in V$ is associated with node feature $v_u$, and each interaction $(u, j, t)$ has edge feature $e^t_{u,v} \in R^{d_e}$, where $d_v$ and $d_e$ denote the dimensions of the node and link feature respectively.

\subsection{Diffusion Model}\label{sec:sde_diffusion}

The diffusion model encompasses both forward and reverse processes.

\noindent \textbf{Forward process.} 
In general, given an input data point $\vx_0$ drawn from the distribution $q(\vx_0)$, the forward process involves gradually introducing Gaussian noise to $\vx_0$, generating a sequence of increasingly noisy variables $\vx_1, \vx_2, \dots, \vx_N$ in a Markov chain. The final noisy output, $\vx_N$, follows a Gaussian distribution $\mathcal{N}(\vzero, \mI)$ and carries no discernible information about the original data point. Specifically, the transition from one point to the next is determined by a conditional probability $q(\vx_\diffusionInd | \vx_{\diffusionInd-1}) = \mathcal{N}(\bm{x}_n; \sqrt{1-\beta_\diffusionInd} \vx_{\diffusionInd-1}, \beta_\diffusionInd \mI)$, where $\beta_n\in (0,1)$ controls the scale of noise added at step $n$.

\noindent \textbf{Reverse process.} 
The reverse process reverses the effects of the forward process by learning to eliminate the added noise and tries to gradually reconstruct the original data $x_0$ via sampling from $x_N$ by learning a neural network $f_\theta$.

\noindent \textbf{Inference.} 
Once trained, the diffusion model can produce new data by sampling a point from the final distribution $\vx_\diffusionLast \sim \mathcal{N}(\vzero, \mI)$ and then iteratively denoising it using the aforementioned model $\vx_\diffusionLast \mapsto \vx_{\diffusionLast -1} \mapsto \dots \mapsto \vx_0$ to obtain a sample from the data distribution.


\begin{figure*}[t]

\centering
    \includegraphics[width=1.0\linewidth]{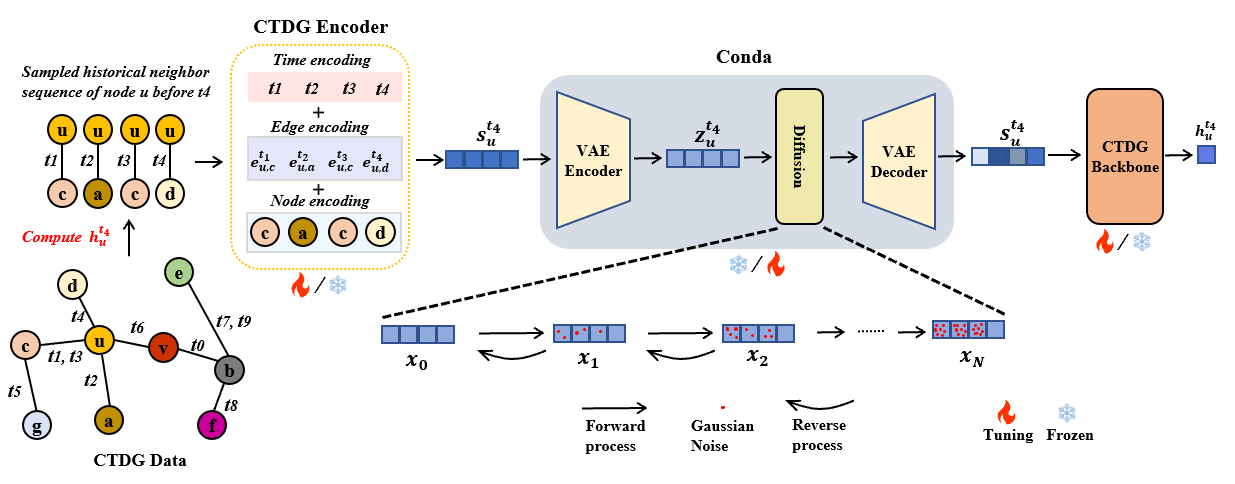}

\caption{The Alternating Training Process of Conda and the CTDG model. While the Conda module is in the training phase, the other modules are frozen. Conversely, when the Conda module is frozen, other modules are in the training phase}\label{fig:model}
\vspace{-10pt}
\end{figure*}

\section{METHODOLOGY}\label{sec:diffusion}
Existing structure-oriented GDA methods like MeTA augment the CTDGs by modifying the initial interactions through edge addition/deletion and time perturbation. However, these methods introduce coarse-grained augmentations and substantially alter the original transition patterns within CTDGs. Conversely, simply introducing noise to either the raw or hidden feature space often lacks theoretical bound. In this section, we introduce a novel fine-grained GDA model based on a conditional diffusion model and establish robust theoretical guarantees.

\subsection{CTDG model}
As mentioned above, the paradigm of the CTDG model $\xi$ can be divided into two parts: the encoder module and the backbone module. Mathematically, given an interaction $(u, v, t)$ and historical interactions before timestamp $t$, the computation flow unfolds as follows:
\begin{equation}
\bm{s}_u^t=enc(\{v_{w_l} \| e_{\left(u, w_l\right)}^{t_l} \| t_l\}) ,w_l \in \mathcal{N}_{<t}(u)
\end{equation}
where $\bm{s}_u^t \in R^{L \times D}$ denotes the historical neighbor embedding sequence, $D$ represents the embedding dimension. $\mathcal{N}_{<t}(u)$ denotes the sampled set of neighbors that interacted with node $u$ before timestamps $t$, $w_l \in \mathcal{N}_{<t}(u)$ is the $l$-th neighbor, and $\|$ denotes the concatenation operation. $enc(\cdot)$ represents the general encoder module of CTDG model.
\begin{equation}
\bm{h}_u^t=backbone(\bm{s}_u^t)
\end{equation}
$\bm{h}_u^t \in R^{D}$ denotes the representation of node $u$ at timestamp $t$. $backbone(\cdot)$ represents the backbone of CTDG model.

\subsection{Conda}

Due to the extensive resource requirement of the diffusion process, to reduce the costs, we first utilize a VAE encoder to conduct dimension compression and then conduct diffusion processes in the latent space. 

\textbf{VAE Encoder.} Given the historical neighbor embedding sequence $\bm{s}$ of any node \footnote{For brevity, we omit the subscript node $u$ and superscript timestamp $t$ in $\bm{s}$ for $\bm{s}_u^t$ unless necessary to avoid ambiguity.}, computed by the CTDG encoder module, we use a variational encoder parameterized by $\phi$ to compress $\bm{s}$ to a low-dimensional vector $\bm{z} \in R^{L \times d}$, where $d<<D$, the VAE encoder $\phi$ predicts $\bm{\mu}_{\phi}$ and $\sigma^2_{\phi}\bm{I}$ as the mean and covariance of the variational distribution $q_{\phi}(\bm{z}|\bm{s})=\mathcal{N}(\bm{z}; \bm{\mu}_{\phi}(\bm{s}),\sigma_{\phi}^2(\bm{s})\bm{I})$.  

\textbf{Forward Process with Partial Noising.} It is worth noting that, different from the conventional diffusion model that corrupts the whole variable without distinction, we conduct partial noising in order to control the magnitude of data augmentation and prepare for the reverse process with conditional denoising. Specifically, given the low-dimensional vector $\bm{z}=[z_{w_1},z_{w_2},...,z_{w_L}]$, we can set $\bm{x}_0=\bm{z}$ as the initial state. Then we divide $\bm{x}_{0}$ into two part: diffused part and conditional part, $\bm{x}_{0}^{diff}\in R^{diff \times d}$ and $\bm{x}_{0}^{cond} \in R^{cond \times d}$, where $diff+cond=L$ and $\bm{x}_{0} = \bm{x}_{0}^{diff}||\bm{x}_{0}^{conda}$. In the forward process, we only add noise on $\bm{x}_{0}^{diff}$. Then the forward process is parameterized by
\begin{equation}\label{eq:q_xt_given_xn-1}
    q(\bm{x}_n^{diff}|\bm{x}_{n-1}^{diff}) = \mathcal{N}(\bm{x}_n^{diff}; \sqrt{1-\beta_n}\bm{x}_{n-1}^{diff}, \beta_n\bm{I}),
\end{equation}
where $\beta_n\in(0,1)$ controls the Gaussian noise scales added at each step $n$. Since the transition kernel is Gaussian, the value at any step $\diffusionInd$ can be sampled directly from $\vx_0$ in practice. Let $\alpha_\diffusionInd = 1 - \beta_\diffusionInd$ and $\bar{\alpha}_\diffusionInd = \prod_{n'=1}^\diffusionInd \alpha_n'$, then we can write:
\begin{align}\label{eq:q_xi_x0}
    q(\vx_\diffusionInd^{diff} | \vx_0^{diff}) = \mathcal{N}(\bm{x}_n^{diff}; \sqrt{\bar{\alpha}_\diffusionInd} \vx_0^{diff}, (1 - \bar{\alpha}_\diffusionInd) \mI).
\end{align}
where we can reparameterize $\bm{x}_n=\sqrt{\bar{\alpha}_n}\bm{x}_0+\sqrt{1-\bar{\alpha}_n}\bm{\epsilon}$ with $\bm{\epsilon}\sim\mathcal{N}(\bm{0},\bm{I})$.
To regulate the added noises in $\bm{x}_{1:N}$, follow~\cite{wang2023diffusion}, we ultilize the linear noise schedule for $1-\bar{\alpha}_n$:
\begin{equation}
    1 - \bar{\alpha}_n = k\cdot\left[\alpha_{\min} + \dfrac{n-1}{N-1}(\alpha_{\max} - \alpha_{\min})\right],\quad n\in\{1,\dots,N\}
    \label{eq:noise_schedule}
\end{equation}
where a hyper-parameter $k\in\left[0,1\right]$ controls the noise scales, and two hyper-parameters $\alpha_{\min}<\alpha_{\max}\in(0,1)$ indicating the upper and lower bounds of the added noises.

\textbf{Reverse Process with Conditional Denoising.}
The reverse process is used to reconstruct the original $\bm{x}_0$ by denoising $\bm{x}_N$. With the partial nosing strategy adopted in the forward process, we can naturally employ the part without noise as the conditional input when denoising. 

Starting from $\bm{x}_N$, the reverse process is parameterized by the denoising transition step:
\begin{equation}
    p_\theta(\bm{x}_{n-1}|\bm{x}_n)=\mathcal{N}(\bm{x}_{n-1};\bm{\mu}_\theta(\bm{x}_n,n),\bm{\sigma}_\theta(\bm{x}_n,n))
    \label{eq:p_nheta}
\end{equation}
where $\bm{\mu}_\theta(\bm{x}_n,n)$ and $\bm{\sigma}_\theta(\bm{x}_n,n)$ are  parameterization of the predicted mean and standard deviation outputted by any neural networks $f_\theta$. Then the whole reverse process can be written as follows:
\begin{equation}
    p_\theta(x_{N:0})=p(x_N)\prod_{n=1}^\diffusionInd p_\theta(\bm{x}_{n-1}|\bm{x}_n)
\end{equation}
The reconstructed output is denoted as $\hat{\bm{x}_0}$.

\textbf{VAE Decoder.} 
To keep the notations consistent, we set $\hat{\bm{z}}=\hat{\bm{x}_0}$, $\hat{\bm{z}}$ is then fed into the VAE decoder parameterized by $\psi$ to predict ${\bm{s}}$ via $p_{\psi}({\bm{s}}|\hat{\bm{x}_0})$. 

\subsection{Optimization and Alternating Training}
In this section, we first present the optimization objective for the Conda and the CTDG model, respectively. Then we introduce the training and inference process of CTDG model with conda.

Although our ultimate goal is to learn a CTDG model $\xi$, but we also have to learn the parameters of the VAE encoder $q_{\phi}(\bm{z}|\bm{s})$, the conditional diffusion model $\log p_\theta(\bm{x}_0)$ and the VAE decoder $p_{\psi}(\bm{s}|\hat{\bm{x_0}})$ for providing positive augmentation to the CTDG model.

\noindent\textbf{CTDG model.} For training the CTDG model $\xi$, the Loss function $\mathcal{L}_{ctdg}$ depends on the downstream task. For example, if the downstream task is link prediction, then the loss function is binary cross-entropy.

\noindent\textbf{Variational Auto Encoder.} The $q_{\phi}(\bm{z}|\bm{s})$ and  $p_{\psi}(\bm{s}|\hat{\bm{x_0}})$ jointly constitute a VAE that bridges the embedding space and the latent space. We optimize the VAE by directly maximizing the ELBO:
\begin{equation}
\small
\begin{aligned}
\mathcal{L}_{vae}(\bm{s};\phi,\psi)=&{\mathbb{E}_{q_{\bm{\phi}}(\vz|\vs)}\left[\log\frac{p(\vs, \vz)}{q_{\bm{\phi}}(\vz|\vs)}\right]}\\
&= {\mathbb{E}_{q_{\bm{\phi}}(\vz|\vs)}\left[\log\frac{p_{\bm{\psi}}(\vs|\vz)p(\vz)}{q_{\bm{\phi}}(\vz|\vs)}\right]}    \\
&={\mathbb{E}_{q_{\bm{\phi}}(\vz|\vs)}\left[\log p_{\bm{\psi}}(\vs|\vz)\right] + \mathbb{E}_{q_{\bm{\phi}}(\vz|\vs)}\left[\log\frac{p(\vz)}{q_{\bm{\phi}}(\vz|\vs)}\right]}         \\
&= \underbrace{{\mathbb{E}_{q_{\bm{\phi}}(\vz|\vs)}\left[\log p_{\bm{\psi}}(\vs|\vz)\right]}}_{\text{reconstruction term}} - \underbrace{D_{\text{KL}}{q_{\bm{\phi}}(\vz|\vs)}{p(\vz)}}_{\text{prior matching term}}\\
& \ge{\mathbb{E}_{q_{\bm{\phi}}(\vz|\vs)}\left[\log p_{\bm{\psi}}(\vs|\vz)\right]}\label{eq:vae_elbo}
\end{aligned}
\end{equation}
where the first term measures the reconstruction likelihood of the decoder from variational distribution, the second term measures how similar the learned variational distribution is to a prior belief held over latent variables. Maximizing the ELBO is thus equivalent to maximizing its first term and minimizing its second term. Since the KL divergence term of the ELBO can be computed analytically, and the reconstruction term can be approximated using a Monte Carlo estimate:
\begin{equation}
\small
\begin{aligned}
\argmax_{\bm{\phi}, \bm{\psi}} \mathbb{E}_{q_{\bm{\phi}}(\vz|\vs)}\left[\log p_{\bm{\psi}}(\vs|\vz)\right] - D_{\text{KL}}({q_{\bm{\phi}}(\vz|\vs)}{p(\vz)}) \\
\approx\argmax_{\bm{\phi}, \bm{\psi}} \sum_{m=1}^{M}\log p_{\bm{\psi}}(\vs|\vz^{(m)}) - D_{\text{KL}}({q_{\bm{\phi}}(\vz|\vs)}{p(\vz)})
\end{aligned}
\end{equation}
where latents $\{\vz^{(m)}\}_{m=1}^M$ are sampled from $q_{\bm{\phi}}(\vz|\vs)$, for every observation $\vs$ in the traning sample.

\textbf{Conditional diffusion model.} To optimize the conditional diffusion model $\theta$, the training objective is to use the Variational Lower Bound (VLB) to optimize the log-likelihood of $\bm{x}_0$:
\begin{equation}\label{eq:diff_ELBO}
\small
\begin{aligned}
\mathcal{L}_{VLB}(\bm{x}_0; \theta)&=\log \mathbb{E}[p_\theta(\bm{x}_0)]\\
& =\log\mathbb{E}[\int p_\theta(\bm{x}_{0:N})\mathrm{d}\bm{x}_{1:N}] \\
&\leq\log\mathbb{E}_{q(\bm{x}_{1:N}|\bm{x}_0)}[\underbrace{\frac{ q(\mathbf{x}_N|\mathbf{x}_0)}{p_{\theta}(\mathbf{x}_N)}}_{\mathcal{L}_N}  + \sum_{n=2}^N \underbrace{\log{\frac{ q(\mathbf{x}_{n-1}|\mathbf{x}_0,\mathbf{x}_n)}{p_{\theta}(\mathbf{x}_{n-1}|\mathbf{x}_n)}}}_{\mathcal{L}_{n-1}}]
\end{aligned} 
\end{equation}
 

Next we will provide detailed  to show how we estimate VLB. The term $\mathcal{L}_{n-1}$ makes $p_\theta(\bm{x}_{n-1}|\bm{x}_n)$ to approximate the tractable distribution $q(\bm{x}_{n-1}|\bm{x}_0,\bm{x}_n)$. Through Bayes rules, 
we can derive the probability of any intermediate value $\vx_{\diffusionInd-1}$ given its successor $\vx_\diffusionInd$ and initial $\vx_0$ as:
\begin{equation}
\begin{aligned}\label{eq:q_posterior}
    q(\vx_{\diffusionInd-1} | \vx_\diffusionInd, \vx_0) = q(\vx_{\diffusionInd} | \vx_{\diffusionInd-1}, \vx_0)\frac{q(\vx_{\diffusionInd-1}|\vx_0)}{q(\vx_{\diffusionInd}|\vx_0)}
\end{aligned}
\end{equation}

\begin{equation}
\begin{aligned}\label{eq:q_posterior}
    q(\vx_{\diffusionInd-1} | \vx_\diffusionInd, \vx_0) = \mathcal{N}(\bm{x}_{n-1}; \tilde{\vmu}_\diffusionInd, \tilde\beta_\diffusionInd \mI)
\end{aligned}
\end{equation}

\begin{equation}
\begin{aligned}
    \text{where\quad} &\tilde{\mu}_\diffusionInd(\vx_n,\vx_0) = \frac{\sqrt{\alpha_\diffusionInd} (1 - \bar\alpha_{\diffusionInd-1})}{1 - \bar\alpha_\diffusionInd} \vx_\diffusionInd +\frac{\sqrt{\bar{\alpha}_{\diffusionInd-1}} \beta_\diffusionInd}{1 - \bar{\alpha}_\diffusionInd} \vx_0 , &\\
    &\tilde\beta_\diffusionInd = \frac{1 - \bar{\alpha}_{\diffusionInd-1}}{1 - \bar{\alpha}_\diffusionInd} \beta_\diffusionInd . &
\end{aligned}\label{eq:fenjie}
\end{equation}

$\tilde{\mu_n}$ denotes the reparameterized mean of $q(\bm{x}_{n-1}|\bm{x}_0,\bm{x}_n)$. Thereafter, for $1 \leq n \leq N-1$, the the parameterization of $\mathcal{L}_{VLB}$ at step $n$ can be calculated by pushing $\bm{\mu}_\theta(\bm{x}_n,n)$ to be close to $\tilde{\bm{\mu}}_n(\vx_n,\vx_0)$. Then, we can similarly factorize $\bm{\mu}_\theta(\bm{x}_n,n)$ via
\begin{equation}
\label{eq:mu_nheta}
    \bm{\mu}_\theta(\bm{x}_n,n) = \dfrac{\sqrt{\alpha_n}(1-\bar{\alpha}_{n-1})}{1-\bar{\alpha}_n}\bm{x}_n+\dfrac{\sqrt{\bar{\alpha}_{n-1}}(1-\alpha_n)}{1-\bar{\alpha}_n}{\bm{f}}_{\theta}(\bm{x}_n,n)
\end{equation}
where $f_{\theta}(\bm{x}_n,n)$ is the predicted $\bm{x}_0$ based on $\bm{x}_n$ and diffusion step $n$. And the $\mathcal{L}_{VLB}$ at step $n$ can be formula as :

\begin{equation}
\small
\begin{aligned}
\mathcal{L}_{VLB_n}
&=\mathbb{E}_{\mathbf{x}_0}\left[\log{\frac{ q(\mathbf{x}_{n}|\mathbf{x}_0,\mathbf{x}_{n+1})}{p_{\theta}(\mathbf{x}_{n}|\mathbf{x}_{n+1})}}\right]\\
&= \mathbb{E}_{\mathbf{x}_0}\left[\frac{1}{2||\sigma_{\theta}||^2}|| \tilde{\mu_n}(\mathbf{x}_n,\mathbf{x}_0)-\mu_{\theta}(\mathbf{x}_n, n)||^2 \right]\\
& =\mathbb{E}_{\mathbf{x}_0}[\frac{1}{2||\sigma_{\theta}||^2}||\;\frac{\sqrt{\alpha_\diffusionInd} (1 - \bar\alpha_{\diffusionInd-1})}{1 - \bar\alpha_\diffusionInd}\mathbf{x}_n+\frac{\sqrt{\bar{\alpha}_{\diffusionInd-1}} \beta_\diffusionInd}{1 - \bar{\alpha}_{\diffusionInd}}\mathbf{x}_0-\\
&(\frac{\sqrt{\alpha_\diffusionInd} (1 - \bar\alpha_{\diffusionInd-1})}{1 - \bar\alpha_\diffusionInd}\mathbf{x}_n + \frac{\sqrt{\bar{\alpha}_{\diffusionInd-1}} \beta_\diffusionInd}{1 -\bar{\alpha}_\diffusionInd}f_{\theta}(\mathbf{x}_n, n))||^2]\\
&=\frac{\frac{\sqrt{\bar{\alpha}_{\diffusionInd-1}} \beta_\diffusionInd}{1 - \bar{\alpha}_\diffusionInd}}{2||\sigma_{\theta}||^2}\mathbb{E}_{\mathbf{x}_0}[||\mathbf{x}_0-f_{\theta}(\mathbf{x}_n, n)||^2],  \\
\end{aligned}
\end{equation}
In practice, to keep training stability and simplify the calculation, we following the previous work~\cite{wang2023diffusion}, ignore the learning of $\bm{\sigma}_\theta(\bm{x}_n,n)$ in $p_\theta(\bm{x}_{n-1}|\bm{x}_n)$ of Eq. (\ref{eq:p_nheta}) and directly set $\bm{\sigma}_\theta(\bm{x}_n,n)=\tilde\beta_\diffusionInd$.Then the optimization of training loss can be further simplified as:

\begin{equation}
\small
\begin{aligned}
\min\;\mathcal{L}_{\text{VLB}}(\bm{x}_0; \theta)=&\min\left[
||\tilde{\mu}(\mathbf{x}_N)||^2+ \sum_{n=2}^N||\mathbf{x}_0-f_{\theta}(\mathbf{x}_n, n)||^2 \right]\\ 
\rightarrow& \min\left[
\sum_{n=2}^N||\mathbf{x}_0-f_{\theta}(\mathbf{x}_n, n)||^2 \right]\\
\end{aligned}
\end{equation}

\noindent\textbf{Optimization of Conda.}

In conclusion, we combine and minimize the loss of the conditional diffusion model and VAE via $ \mathcal{L}_{VLB}(\bm{x}_0;\theta)+ \lambda\cdot\mathcal{L}_{vae}(\bm{s};\phi,\psi)$ to optimize Conda, where the hyper-parameter $\lambda$ ensures the two terms in the same magnitude.

\noindent\textbf{Alternating training.} Unlike the common diffusion models that are trained for the direct generation of raw graph data through pre-training, Conda requires the historical neighbor sequence embeddings of nodes obtained through the CTDG encoder before performing the diffusion process. Therefore, we utilize alternative training method to alternatively train the CTDG model and Conda. 
Here we briefly describe the training process. Initially, the CTDG model $\xi$ is trained by minimizing the $\mathcal{L}_{ctdg}$ for $R_ctdg$ rounds. Then we insert Conda into the intermediate layer of the CTDG model and train the $\theta, \phi, \psi$ according to $\mathcal{L}_{VLB}(\bm{x}_0;\theta)+ \lambda\cdot\mathcal{L}_{vae}(\bm{s};\phi,\psi)$ with the $\xi$ frozen for $R_conda$ rounds. Next, we train the CTDG model $\xi$ again for $R_ctdg$ rounds with the $\theta, \phi, \psi$ frozen. At this point, Conda is in the inference phase, used to generate augmented historical neighbor embeddings via the reverse process. The above process will be repeated several times.


\begin{table*}[t]
\centering
\renewcommand\arraystretch{1.5}
\setlength{\tabcolsep}{0.6mm}{\footnotesize

\begin{tabular}{c|ccccccccccccc}
\hline
                \multirow{2}{*}{Method} & \multicolumn{2}{c}{Wiki\_0.1} & \multicolumn{2}{c}{Reddit\_0.1}  & \multicolumn{2}{c}{MOOC\_0.1} & \multicolumn{2}{c}{UCI\_0.1}  & \multicolumn{2}{c}{LastFM\_0.1} & \multicolumn{2}{c}{Social.Evo\_0.1}  \\ \cline{2-13} 
                                & AP   & A-R   & AP    & A-R  & AP  & A-R & AP   & A-R   & AP   & A-R    & AP   & A-R    \\ \hline
                JODIE &85.57±1.77& 89.16±0.51 &94.01±1.02&94.82±0.81&70.57±0.15&75.44±0.09&84.43±0.62&87.44±0.25&64.94±1.22&66.17±1.14&77.11±0.90&82.86±0.76\\
                JODIE+Conda &\textbf{88.63±0.93}&\textbf{90.52±0.40}&\textbf{94.55±0.82}&\textbf{94.98±0.65}&\textbf{72.11±0.47}&\textbf{75.79±0.42}&\textbf{84.57±0.70}&\textbf{87.71±0.59}&\textbf{65.09±1.24}&\textbf{66.25±1.17}&\textbf{78.74±0.71}&\textbf{83.46±0.66}\\\hline
                   DyRep &93.86±0.06 &93.37±0.09 &94.24±0.75 & 94.64±0.29 &71.45±1.36 &76.81±1.33 &69.19±0.90 &73.70±0.71&65.86±0.54&66.61±0.59&77.57±0.64&82.99±0.60 \\
                   DyRep+Conda &\textbf{94.05±0.13}&\textbf{93.89±0.22}&\textbf{94.84±0.64}&\textbf{94.98±0.22}&\textbf{72.88±1.12}&\textbf{77.02±1.07}&\textbf{69.47±0.94}&\textbf{73.91±0.85}&\textbf{66.37±0.51}&\textbf{66.94±0.48}&\textbf{78.97±0.43}&\textbf{83.62±0.37}\\\hline
                   TGAT &93.16±0.29 &93.25±0.30 &94.02±0.12  &94.01±0.13 &74.44±1.27 &74.13±1.92 &73.54±1.05&73.01±1.22&69.93±0.30&70.42±0.33&80.03±0.46&85.09±0.41\\
                   TGAT+Conda &\textbf{94.02±0.27}&\textbf{94.11±0.29}&\textbf{94.88±0.10}&\textbf{94.90±0.09}&\textbf{76.15±1.43}&\textbf{76.97±1.85}&\textbf{73.99±1.10}&\textbf{73.74±1.18}&\textbf{69.94±0.28}&\textbf{70.48±0.41}&\textbf{81.79±0.60}&\textbf{85.83±0.57}\\\hline
                   TGN   &93.01±1.22 &92.77±1.05 &94.94±0.09&94.23±0.12& 77.30±0.50&75.62±0.72& 88.00±0.99& 87.73±1.14&73.61±0.74&72.75±0.68&78.79±0.64&78.60±0.48 \\
                   TGN+Conda &\textbf{93.47±1.16}&\textbf{92.93±1.09}&\textbf{95.50±0.14}&\textbf{95.06±0.18}&\textbf{78.63±0.75}&\textbf{76.52±0.81}&\textbf{88.22±0.92}&\textbf{87.79±1.06}&\textbf{74.24±0.80}&\textbf{73.31±0.75}&\textbf{80.20±0.57}&\textbf{80.03±0.48}\\\hline
                   TCL   &93.48±0.27&92.98±0.23& 95.05±0.11&94.92±0.12&74.12±0.74&78.10±0.65&85.14±1.54&84.87±1.50&60.07±0.15&58.97±0.17&83.57±1.22&85.41±1.04\\
                   TCL+Conda &\textbf{94.04±0.20}&\textbf{93.75±0.17}&\textbf{95.96±0.08}&\textbf{95.79±0.07}&\textbf{76.31±1.20}&\textbf{79.55±1.13}&\textbf{85.79±1.37}&\textbf{85.40±1.28}&\textbf{60.15±0.44}&\textbf{59.21±0.45}&\textbf{86.09±1.25}&\textbf{88.45±1.17}\\\hline
                  GraphMixer  &94.02±0.13&93.73±0.12&94.93±0.06&94.62±0.06& 74.15±1.92& 78.07±1.67 &90.10±1.51 &89.83±1.42 &71.11±0.07&70.33±0.09&82.40±1.09&86.24±0.92 \\ 
                  GraphMixer+Conda &\textbf{94.61±0.19}&\textbf{94.26±0.20}&\textbf{95.17±0.09}&\textbf{94.85±0.08}&\textbf{75.24±1.66}&\textbf{78.79±1.42}&\textbf{90.33±1.70}&\textbf{90.00±1.64}&\textbf{71.70±0.18}&\textbf{70.85±0.19}&\textbf{83.89±0.84}&\textbf{88.31±0.77} \\\hline
                  DyGFormer  &95.74±0.11&95.54±0.09&96.01±0.08&96.00±0.07&75.47±0.96& 77.49±0.81 &91.02±0.85 &90.74±0.67&77.86±0.14&77.01±0.10&84.25±0.73&87.64±0.62\\ 
                  DyGFormer+Conda &\textbf{96.68±0.14}&\textbf{96.32±0.13}&\textbf{96.68±0.14}&\textbf{96.55±0.12}&\textbf{76.12±1.41}&\textbf{77.98±1.33}&\textbf{91.09±1.00}&\textbf{90.97±0.92}&\textbf{78.84±0.35}&\textbf{78.11±0.27}&\textbf{85.64±0.58}&\textbf{88.78±0.40}\\\hline

\end{tabular}}
\caption{Experiments results on dataset with 0.1 ratio of train set}\label{tab:0.1}
\vspace{-5pt}
\end{table*}

\begin{table*}[t]
\centering
\renewcommand\arraystretch{1.5}
\setlength{\tabcolsep}{0.6mm}{\footnotesize

\begin{tabular}{c|ccccccccccccc}
\hline
                \multirow{2}{*}{Method} & \multicolumn{2}{c}{Wiki\_0.3} & \multicolumn{2}{c}{Reddit\_0.3}  & \multicolumn{2}{c}{MOOC\_0.3} & \multicolumn{2}{c}{UCI\_0.3}  & \multicolumn{2}{c}{LastFM\_0.3} & \multicolumn{2}{c}{Social.Evo\_0.3}  \\ \cline{2-13} 
                                & AP   & A-R   & AP    & A-R  & AP  & A-R & AP   & A-R   & AP   & A-R    & AP   & A-R    \\ \hline
                JODIE &90.23±0.41&91.47±0.35&95.63±0.37&95.72±0.33&72.28±0.64&75.41±0.59&85.17±0.88&87.45±0.27&65.62±1.54&67.04±1.44&78.65±0.58&83.43±0.54\\
                JODIE+Conda &\textbf{91.54±0.55}&\textbf{92.01±0.48}&\textbf{96.55±0.36}&\textbf{96.60±0.28}&\textbf{73.75±0.85}&\textbf{75.97±0.54}&\textbf{85.67±1.33}&\textbf{87.79±1.01}&\textbf{66.13±1.87}&\textbf{67.75±1.60}&\textbf{79.42±0.78}&\textbf{84.05±0.70}\\\hline
                   DyRep  &94.11±0.16&93.96±0.13&95.78±0.34&95.90±0.15&73.46±1.17&\textbf{78.04±1.11}&\textbf{71.40±0.75}&\textbf{75.71} ±1.57&67.41±0.86& 68.02±0.91&77.12±0.50&78.51±0.48\\
                   DyRep+Conda &\textbf{94.42±0.33}&\textbf{94.15±0.29}&\textbf{96.58±0.28}&\textbf{96.63±0.21}&\textbf{74.23±1.07}&77.69±1.05&70.33±0.27&74.31±1.40&\textbf{67.94±0.75}&\textbf{68.35±0.92}&\textbf{80.65±0.63}&\textbf{81.00±0.57}\\\hline
                   TGAT &94.33±0.25&94.10±0.20&95.96±0.08&96.02±0.06&78.31±0.97&80.11±0.78&71.04±0.43&72.43±0.63&70.65±0.38&70.22±0.40&82.59±0.78&86.97±0.74\\
                   TGAT+Conda &\textbf{94.99±0.20}&\textbf{94.58±0.19}&\textbf{96.58±0.10}&\textbf{96.77±0.09}&\textbf{80.04±1.23}&\textbf{80.85±1.10}&\textbf{71.44±0.50}&\textbf{72.69±0.65}&\textbf{71.64±0.27}&\textbf{71.08±0.31}&\textbf{83.22±0.80}&\textbf{87.41±0.79}\\\hline
                   TGN    &95.90±0.32&95.66±0.30&96.25±0.13&96.12±0.10&82.31±1.86&81.49±2.03&\textbf{88.96±0.57}&88.02±0.51&75.68±1.74&74.95±1.55&81.85±0.32&85.73±0.28\\
                   TGN+Conda &\textbf{96.52±0.37}&\textbf{96.24±0.35}&\textbf{96.97±0.16}&\textbf{96.78±0.15}&\textbf{82.55±1.65}&\textbf{81.63±1.79}&88.92±0.68&\textbf{88.09±0.64}&\textbf{76.61±1.66}&\textbf{75.13±1.47}&\textbf{84.20±0.44}&\textbf{88.36±0.35}\\\hline
                   TCL    &95.75±0.10&95.01±0.11&96.50±0.06&96.42± 0.05&80.20±0.14&81.98±0.21&88.62±0.24&87.84±0.19&62.23±0.88&59.83±0.75&91.57±0.25&93.64±0.22\\
                   TCL+Conda &\textbf{96.23±0.20}&\textbf{95.58±0.24}&\textbf{97.02±0.12}&\textbf{96.87±0.11}&\textbf{80.47±0.35}&\textbf{82.24±0.32}&\textbf{88.69±0.30}&\textbf{87.93±0.27}&\textbf{62.79±1.14}&\textbf{60.56±1.02}&\textbf{91.88±0.23}&\textbf{93.95±0.19}\\\hline
                  GraphMixer  &95.72±0.05&95.47±0.02&96.46±0.03&96.12±0.01&79.72±0.37&82.22±0.26&91.14±1.03 &90.68±0.94&72.17±0.14&72.05±0.13&87.86±0.63&90.73±0.58 \\
                  GraphMixer+Conda &\textbf{95.99±0.05}&\textbf{95.75±0.03}&\textbf{96.89±0.04}&\textbf{96.65±0.03}&\textbf{80.01±0.58}&\textbf{82.52±0.51}&\textbf{91.84±1.00}&\textbf{90.87±0.93}&\textbf{72.62±0.16}&\textbf{72.17±0.16}&\textbf{89.05±0.57}&\textbf{91.89±0.53} \\\hline
                  DyGFormer  &96.50±0.06&96.36±0.05&97.14±0.03&97.07±0.03&79.42±0.49&83.09±0.31& 92.87±0.62&92.45±0.53&82.57±0.48&81.79±0.40&88.91±0.47&91.54±0.43\\ 
                  DyGFormer+Conda &\textbf{97.11±0.09}&\textbf{96.96±0.10}&\textbf{97.45±0.04}&\textbf{97.31±0.04}&\textbf{81.00±0.83}&\textbf{83.97±0.53}&\textbf{92.94±0.48}&\textbf{92.50±0.40}&\textbf{83.17±0.36}&\textbf{82.31±0.32}&\textbf{90.02±0.53}&\textbf{92.79±0.50}\\\hline

\end{tabular}}
\caption{Experiments results on the dataset with 0.3 ratio of train set}\label{tab:0.3}
\vspace{-5pt}
\end{table*}

\begin{table*}[t]
\centering
\renewcommand\arraystretch{1.5}
\setlength{\tabcolsep}{0.6mm}{\footnotesize

\begin{tabular}{c|ccccccccccccc}
\hline
                \multirow{2}{*}{Method} & \multicolumn{2}{c}{Wiki\_0.3} & \multicolumn{2}{c}{Reddit\_0.3}  & \multicolumn{2}{c}{MOOC\_0.3} & \multicolumn{2}{c}{UCI\_0.3}  & \multicolumn{2}{c}{LastFM\_0.3} & \multicolumn{2}{c}{Social.Evo\_0.3}  \\ \cline{2-13} 
                                & AP   & A-R   & AP    & A-R  & AP  & A-R & AP   & A-R   & AP   & A-R    & AP   & A-R    \\ \hline

                   TGN    &95.90±0.32&95.66±0.30&96.25±0.13&96.12±0.10&82.31±1.86&81.49±2.03&\textbf{88.96±0.57}&88.02±0.51&75.68±1.74&74.95±1.55&81.85±0.32&85.73±0.28\\
                   TGN+DropEdge &96.11±0.69&95.69±0.92&96.79±0.86&96.72±0.72&82.46±1.91&81.54±1.54&88.93±0.90& 88.09±0.89&75.91±1.92&75.02±1.50&82.15±0.51&86.20±0.48\\
                   TGN+DropNode &96.03±0.73& 95.97±0.76& 96.87±0.87& 96.63±0.63& 82.35±1.35&81.62±1.62 &88.96±0.95& 88.05±0.50&75.77±1.76&75.00±1.00& 82.31±0.67&86.55±0.60  \\
                   TGN+MeTA &96.20±0.91& 95.97±0.97& 96.75±0.75&96.61±0.61&82.31±1.31&81.60±1.60& 88.92±0.92 &88.05±0.54& 76.47±1.47&75.12±1.12&83.27±0.63&86.94±0.56 \\
                   TGN+Conda &\textbf{96.52±0.37}&\textbf{96.24±0.35}&\textbf{96.97±0.16}&\textbf{96.78±0.15}&\textbf{82.55±1.65}&\textbf{81.63±1.79}&88.92±0.68&\textbf{88.09±0.64}&\textbf{76.61±1.66}&\textbf{75.13±1.47}&\textbf{84.20±0.44}&\textbf{88.36±0.35}\\\hline
                  GraphMixer  &95.72±0.05&95.47±0.02&96.46±0.03&96.12±0.01&79.72±0.37&82.22±0.26&91.14±1.03 &90.68±0.94&72.17±0.14&72.05±0.13&87.86±0.63&90.73±0.58 \\
                  
                  GraphMixer+DropEdge &95.98±0.27&95.70±0.24&96.63±0.62&96.49±0.49&79.74±0.73& 82.48±0.47& 91.45±0.45&90.83±0.83& 72.31±0.30&72.08±0.18& 88.91±0.90&0.84±0.83\\
                  GraphMixer+DropNode &95.90±0.32& 95.60±0.30& 96.72±0.72&96.32±0.31& 79.75±0.74& 82.42±0.41& 91.47±0.47& 90.74±0.74& 72.25±0.25& 72.17±0.16& 88.76±0.76& 90.88±0.87 \\
                  GraphMixer+MeTA & 95.97±0.97&95.47±0.47&96.59±0.59&96.49±0.48&79.87±0.86& 82.44±0.43& 91.29±0.29& 90.81±0.81& 72.18±0.17& 72.15±0.14& 88.34±0.94& 90.97±0.96\\
                  GraphMixer+Conda &\textbf{95.99±0.05}&\textbf{95.75±0.03}&\textbf{96.89±0.04}&\textbf{96.65±0.03}&\textbf{80.01±0.58}&\textbf{82.52±0.51}&\textbf{91.84±1.00}&\textbf{90.87±0.93}&\textbf{72.62±0.16}&\textbf{72.17±0.16}&\textbf{89.05±0.57}&\textbf{91.89±0.53} \\\hline
                  DyGFormer  &96.50±0.06&96.36±0.05&97.14±0.03&97.07±0.03&79.42±0.49&83.09±0.31& 92.87±0.62&92.45±0.53&82.57±0.48&81.79±0.40&88.91±0.47&91.54±0.43\\ 
                  DyGFormer+DropEdge &96.71±0.24& 96.48±0.21& 97.35±0.14& 97.12±0.12& 80.37±0.63&83.94±0.50& 92.88±0.55&92.46±0.32& 82.84±0.62& 82.17±0.20& 89.09±0.62& 91.70±0.49\\
                  DyGFormer+DropNode  &96.72±0.24& 96.51±0.21& 97.43±0.15& 97.20±0.15& 79.52±0.64&83.86±0.46&92.93±0.19& 92.48±0.58& 82.83±0.64& 82.14±0.32& 89.04±0.84& 92.24±0.51\\
                  DyGFormer+MeTA &96.57±0.33& 96.38±0.28& 97.26±0.79& 97.19±0.62&79.53±1.05&83.40±0.82& 92.60±0.94& 92.49±0.05& 83.09±0.51& 81.94±0.14& 89.93±0.53& 91.64±0.21 \\
                  DyGFormer+Conda &\textbf{97.11±0.09}&\textbf{96.96±0.10}&\textbf{97.45±0.04}&\textbf{97.31±0.04}&\textbf{81.00±0.83}&\textbf{83.97±0.53}&\textbf{92.94±0.48}&\textbf{92.50±0.40}&\textbf{83.17±0.36}&\textbf{82.31±0.32}&\textbf{90.02±0.53}&\textbf{92.79±0.50}\\\hline

\end{tabular}}
\caption{Performance comparison of baseline and baseline with different GDA methods on the dataset with 0.3 ratio of train set}\label{tab:comp_0.3}
\vspace{-5pt}
\end{table*}

\begin{table*}[t]
\centering
\renewcommand\arraystretch{1.5}
\setlength{\tabcolsep}{0.6mm}{\footnotesize

\begin{tabular}{c|ccccccccccccc}
\hline
                \multirow{2}{*}{Method} & \multicolumn{2}{c}{Wiki\_0.1} & \multicolumn{2}{c}{Reddit\_0.1}  & \multicolumn{2}{c}{MOOC\_0.1} & \multicolumn{2}{c}{UCI\_0.1}  & \multicolumn{2}{c}{LastFM\_0.1} & \multicolumn{2}{c}{Social.Evo\_0.1}  \\ \cline{2-13} 
                                & AP   & A-R   & AP    & A-R  & AP  & A-R & AP   & A-R   & AP   & A-R    & AP   & A-R    \\ \hline

                   TGN   &93.01±1.22 &92.77±1.05 &94.94±0.09&94.23±0.12& 77.30±0.50&75.62±0.72& 88.00±0.99& 87.73±1.14&73.61±0.74&72.75±0.68&78.79±0.64&78.60±0.48 \\
                   TGN+DropEdge &93.18±1.40&92.90±1.51& 95.12±0.58& 94.38±0.33& 76.62±1.98& 75.60±1.62& 87.95±1.69& 87.73±1.25& 74.12±1.05& 72.85±1.77& 78.66±1.58& 78.06±1.10\\
                   TGN+DropNode &92.82±1.42& 92.71±1.48& 95.26±0.69& 94.44±0.41&76.90±1.54& 75.93±1.91& 88.04±1.28& 87.77±1.90&73.86±1.69& 73.23±1.28&79.21±1.37& 79.50±1.21 \\
                   TGN+MeTA  &92.96±1.45& 92.85±1.37& 95.20±0.84& 93.83±0.50& 78.03±1.88& 76.38±1.88& 87.93±1.87& 87.73±1.47& 74.15±1.03& 73.11±1.19&79.43±1.31& 78.57±1.08\\
                   TGN+Conda &\textbf{93.47±1.16}&\textbf{92.93±1.09}&\textbf{95.50±0.14}&\textbf{95.06±0.18}&\textbf{78.63±0.75}&\textbf{76.52±0.81}&\textbf{88.22±0.92}&\textbf{87.79±1.06}&\textbf{74.24±0.80}&\textbf{73.31±0.75}&\textbf{80.20±0.57}&\textbf{80.03±0.48}\\\hline
                  GraphMixer  &94.02±0.13&93.73±0.12&94.93±0.06&94.62±0.06& 74.15±1.92& 78.07±1.67 &90.10±1.51 &89.83±1.42 &71.11±0.07&70.33±0.09&82.40±1.09&86.24±0.92 \\ 
                  GraphMixer+DropEdge &94.34±0.71&93.61±0.43&94.97±0.31&94.84±0.24&75.02±1.72&78.06±1.11&90.14±1.37&89.77±1.30&70.75±0.68&70.51±0.40&82.88±1.16&85.90±1.02\\
                  GraphMixer+DropNode  &93.97±0.74&93.89±0.45&94.90±0.39&94.82±0.40&73.54±1.98&78.15±1.06&90.06±1.64&89.61±1.50&71.28±0.73&70.41±0.70&82.67±1.94&86.12±1.82\\
                  GraphMixer+MeTA  &93.72±0.96&94.21±0.72&94.98±0.39&94.58±0.33&74.35±1.55&77.97±1.44&90.11±2.00&89.86±1.64&71.70±0.41&70.75±0.58&82.74±1.88&87.02±1.61\\
                   GraphMixer+Conda &\textbf{94.61±0.19}&\textbf{94.26±0.20}&\textbf{95.17±0.09}&\textbf{94.85±0.08}&\textbf{75.24±1.66}&\textbf{78.79±1.42}&\textbf{90.33±1.70}&\textbf{90.00±1.64}&\textbf{71.70±0.18}&\textbf{70.85±0.19}&\textbf{83.89±0.84}&\textbf{88.31±0.77} \\\hline
                 DyGFormer  &95.74±0.11&95.54±0.09&96.01±0.08&96.00±0.07&75.47±0.96& 77.49±0.81 &91.02±0.85 &90.74±0.67&77.86±0.14&77.01±0.10&84.25±0.73&87.64±0.62\\ 
                  DyGFormer+DropEdge &96.05±0.57&95.68±0.33&96.22±0.42&95.90±0.30&75.92±1.15&77.76±1.98&91.05±1.41&90.87±1.26&76.93±1.30&77.91±1.81&85.18±1.55&88.71±1.54\\
                  DyGFormer+DropNode &96.02±0.61&95.68±0.52&95.54±0.45&95.50±0.36&74.94±1.87&77.20±1.30&90.89±1.88&90.74±1.06&78.20±1.15&77.17±1.63&83.76±1.75&88.16±1.34 \\
                  DyGFormer+MeTA  &96.24±0.58&95.83±0.33&96.05±0.61&96.00±0.49&75.70±1.21&77.64±1.00&90.76±1.64&90.96±1.45&77.96±1.27&77.99±1.25&85.48±1.72&88.59±1.55\\
                  DyGFormer+Conda &\textbf{96.68±0.14}&\textbf{96.32±0.13}&\textbf{96.68±0.14}&\textbf{96.55±0.12}&\textbf{76.12±1.41}&\textbf{77.98±1.33}&\textbf{91.09±1.00}&\textbf{90.97±0.92}&\textbf{78.84±0.35}&\textbf{78.11±0.27}&\textbf{85.64±0.58}&\textbf{88.78±0.40}\\\hline

\end{tabular}}
\caption{Performance comparison of baseline and baseline with different GDA methods on the dataset with 0.1 ratio of train set}\label{tab:comp_0.1}
\vspace{-5pt}
\end{table*}

\section{Experiments}
In this section, we evaluate the performance of our method on link prediction task across various CTDG models. All the experiments are conducted on open CTDG datasets.

\subsection{Experiment settings}
\subsubsection{\textbf{Datasets}} 

We utilize six open CTDG datasets: Wiki, REDDIT, MOOC, LastFM, Social Evo, and UCI in the experiments. The detailed description and the statistics of the datasets are shown in Table~\ref{tab:statTable} in Appendix~\ref{sec:dataset}. 
The sparsity of the graphs is quantified using the density score, calculated as $\frac{2|E|}{|V|(|V| - 1)}$, where $|E|$ and $|V|$ represent the number of links and nodes in the training set, respectively. These datasets are split into three chronological segments for training, validation, and testing with ratios of 10\%-10\%-80\% and 30\%-20\%-50\%. To differentiate the datasets with different splitting ratios, the dataset names are written with suffix 0.1 and 0.3.

\subsubsection{\textbf{Baselines}}
Since Conda is agnostic to model structure, to evaluate the performance of our GDA method, we conduct experiments on several state-of-the-art CTDG models, including JODIE~\citep{JODIE}, DyRep~\citep{DyRep}, TGAT~\citep{TGAT}, TGN~\citep{TGN}, TCL~\citep{wang2021tcl}, GraphMixer~\citep{graphmixer}, DyGFormer~\citep{DyGFormer}. We also combine our method with other data augmentation methods: DropEdge~\cite{rong2020dropedge}, DropNode~\cite{dropnode}, and MeTA~\cite{wang2021adaptive}. Detailed descriptions of these baselines and GDA methods can be found in Appendix~\ref{sec:baselines} and Appendix~\ref{sec:GDAmethods}, respectively.

\subsubsection{\textbf{Evaluation and hyper-parameter settings}} We evaluate Conda on the task of link prediction. As for the evaluation metrics, we follow the previous works~\cite{DyGFormer,wang2021adaptive}, employing Average Precision (AP) and Area Under the Receiver Operating Characteristic Curve (A-R) as the evaluation metrics. We perform the grad search to find the best settings of some critical hyper-parameters. We vary the learning rates of all baselines in $\{1e^{-4},1e^{-3}\}$, the dropout rate of dropout layer in $\{0.0, 0.1, 0.2, 0.3, 0.4, 0.5\}$, the number $L$ of sampled neighbors and the diffusion length $diffusion$ in$\{10,20,32,64,128,256,512\}$ and $\{1, \frac{L}{16},\frac{L}{8},\frac{L}{4},\frac{L}{3}\}$, respectively. The number of diffusion steps $N$ is fixed at 50, respectively. Besides, the noise scale $k$ is tuned in $\{1e^{-5},1e^{-4},1e^{-3},1e^{-2}\}$. Regarding GDA methods used for comparison, we vary the drop rate for Dropedge and Dropnode in $\{0.1, 0.2, 0.3, 0.4, 0.5\}$. As for MeTA, we control the magnitude of the three DA strategies with a unified $p$, vary in $\{0.1, 0.2, 0.3\}$, and follow the setting in its paper. More details are listed in the Appendix.

The configurations of the baselines align with those specified in their respective papers. The model that achieves the highest performance on the validation set is selected for testing. We conduct five runs of each method with different seeds and report the average performance to eliminate deviations. All experiments are performed on a server with 8 NVIDIA A100-SXM4 40GB GPUs.

\subsection{Performance Comparison and Discussion}
In this section, in order to verify the effectiveness of Conda, we integrate it into each baseline across six datasets with different ratios of the train set for the link prediction task. As shown in Table~\ref{tab:0.1} and Table~\ref{tab:0.3}, mean and standard deviations of five runs are reported, and the best results are highlighted in \textbf{bold font}. The experiment results clearly demonstrate that Conda improves the performance of all the baselines with respect to all datasets with different ratios of train sets.

From the results, we can observe that with the ratio of train set decreasing, the performance of each baseline also decreases. Specifically, when the training data is relatively sufficient, all baselines achieve great performance. However, when training data is more limited, the performance of most baselines drops significantly (e.g. JODIE on Wiki\_0.1, all baselines on Reddit\_0.1, MOOC\_0.1 and SocialEvo\_0.1). The possible reason is that the paradigm of CTDG models is to use historical data to obtain target node embeddings. When historical data is limited, the quality of the obtained embeddings cannot be guaranteed. In addition, the data distribution of the testing set could be diverse from the training set. This would lead to the model overfitting the historical data and cannot be generalized to future data. By using Conda, the model's performance improves. It is achieved by utilizing the conditional diffusion model to generate augmented historical neighbor embeddings of the target node during the training of the CTDG model. In Conda, the mechanism of partial noise addition and conditional inputs ensures that the newly generated embeddings are not random. Instead, they closely resemble the embeddings of recently interacted neighbors. Consequently, this guarantees high-quality embeddings of the node's historical neighbors following augmentation.

In addition, we also compare Conda with three GDA methods to show the superiority of our GDA method. Overall, we find that Conda can consistently outperform competing GDA methods. Furthermore, the variance in the results indicates that Conda provides stable improvements in model performance, unlike other GDA methods that rely on random augmentations and thus yield erratic results. This stability is particularly evident in the setting with more sparse training data (e.g. 0.1 train set ratio). By analyzing the experiments on datasets with a 0.3 train set ratio, it's obvious that all GDA methods can improve baselines' performance on most datasets to some extent, but the improvement on UCI\_0.3 is relatively minor compared to SocialEvo\_0.3. The reason may be that SocialEvo has a smaller sparsity and longer interaction sequences than UCI. This phenomenon suggests that training the CTDG model indeed requires sufficient historical interaction data. Moreover, we notice that MeTA improves baselines' performance than DropEdge and DropNode. This might be because MeTA considers the time perturbation, which is crucial for dynamic graph learning. Additionally, unlike Dropedge and DropNode, which essentially remove edges, MeTA maintains or even increases the number of interaction data samples before and after augmentation by simultaneously adding and removing edges. However, due to the combination of multiple augmentation strategies, the results of MeTA also introduce a larger variance, indicating that it is hard to control. 

Next, we analyze the results of the experiment on datasets with a 0.1 train set ratio. Table~\ref{tab:comp_0.1} clearly shows that, apart from our method, which still achieves stable performance improvements, the other three methods frequently resulted in outcomes even worse than the original baseline. The main reason is that at such a low ratio of train sets, the datasets become extremely sparse. At this level, employing random perturbations like edge deletion further reduces the already insufficient data samples and risks removing essential interaction. However, our method maintains stable performance gains by controlling the diffused sequence length. Even though the dataset becomes sparser and the historical neighbors of nodes decrease, by simultaneously reducing the length of the diffused sequence, we still ensure a stable, albeit somewhat reduced, level of improvement compared to the results on datasets with a 0.3 ratio of the train set. The effect of the diffused length on model performance will be analyzed in detail in the following section.

\begin{table}[t]
\renewcommand\arraystretch{1.5}
\setlength{\tabcolsep}{0.1mm}{\footnotesize
\begin{tabular}{cccccccc}
\hline
                         & WIKI\_0.1     & Reddit\_0.1    & MOOC\_0.1     & UCI\_0.1  &LastFM\_0.1          \\\hline
GraphMixer       & 94.02±0.13&94.93±0.06&74.15±1.92& 90.10±1.51&71.11±0.07 \\
+Conda w/o VAE &  94.58±0.17 &95.12±0.08 &74.90±2.17 &90.30±1.67 &71.65±0.18      \\
+Conda w/o diffusion   &93.87±0.25 & 94.80±0.13 &74.77±1.80 & 90.15±1.91&71.43±0.30 \\ 
+Conda       &\textbf{94.61±0.19} &\textbf{95.17±0.09} & \textbf{75.24±1.66} & \textbf{90.33±1.70} &\textbf{71.70±0.18}\\
 \hline
\end{tabular}}
\caption{Experiment results on AP of different variants}\label{tab:ab_module}
\vspace{-10pt}
\end{table}

\begin{table}[t]
\renewcommand\arraystretch{1.5}
\setlength{\tabcolsep}{1.0mm}{\small
\begin{tabular}{cccccccc}
\hline
                         & WIKI\_0.1  & MOOC\_0.1  & WIKI\_0.3  & MOOC\_0.3            \\\hline
GraphMixer      & 94.02±0.13&74.15±1.92&95.72±0.05 &79.72±0.37 \\
+Conda (E2E)    &94.09 ± 0.58 & 73.62± 1.90 &95.49±0.13 & 78.53 ± 0.90\\
+Conda (AT)    &\textbf{94.61±0.19} & \textbf{75.24±1.66} &\textbf{95.99±0.05} &\textbf{80.01±0.58}\\\hline
\end{tabular}}
\caption{AP of different training approaches}\label{tab:AT}
\vspace{-10pt}
\end{table}

\subsection{Ablation analysis}
We conduct an ablation study to assess the contributions of the VAE and diffusion components within the Conda module. The results, summarized in Table~\ref{tab:ab_module}, compare the baseline GraphMixer, add Conda without VAE (+Conda w/o VAE), add Conda without diffusion (+Conda w/o diffusion), and add the full Conda module (+Conda). 

It is obvious that the full Conda module achieves the highest AP scores on all datasets, and consistently outperforms all variants, indicating the importance of both VAE and diffusion components. Removing the diffusion component results in a performance drop, particularly on WIKI\_0.1 (from 94.61 to 93.87), highlighting the diffusion’s role in generating effective latent representation. Similarly, removing the VAE component also decreases AP scores, especially on MOOC\_0.1 (from 75.24 to 74.77). In conclusion, the combination of the VAE and diffusion model results in superior performance, as shown by consistently higher AP scores compared to the ablated variants. This synergy is crucial for optimal model performance.

In addition to the ablation study of the Conda module, we also explore different training approaches to further understand their impact on model performance. As shown in Table~\ref{tab:AT}, we conduct experiments on GraphMixer+Conda with two different training approaches: end-to-end training (E2E) and alternative training (AT). The results indicate that the E2E training approach results in a significant performance decline across all datasets. For example, on the MOOC\_0.1 and MOOC\_0.3, the AP drops from 75.24.61 (AT) to 73.62 (E2E) and from 80.01 (AT) to 78.53 E2E), respectively. This decline can be attributed to conflicting objectives between the Conda module and the CTDG model. The Conda module aims to generate embeddings similar to the original data, while the CTDG model seeks to learn from diverse augmented data close to reality to enhance performance. When integrated into end-to-end training, these conflicting goals prevent the model from optimally achieving both objectives, leading to suboptimal or even diminishing performance.

\begin{figure*}[t]
\centering
    \includegraphics[width=0.31\linewidth]{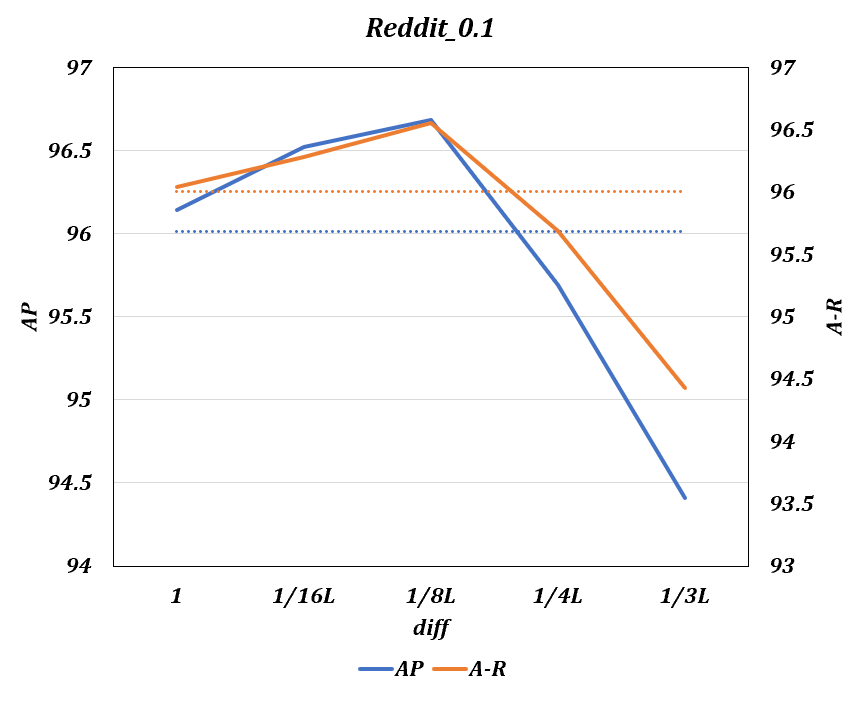}
    \includegraphics[width=0.31\linewidth]{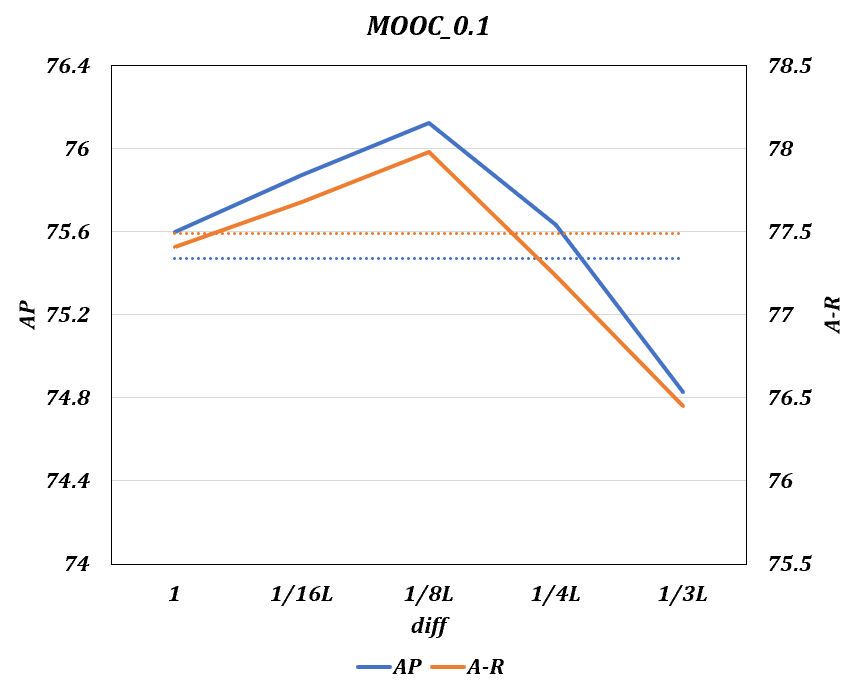}
    \includegraphics[width=0.31\linewidth]{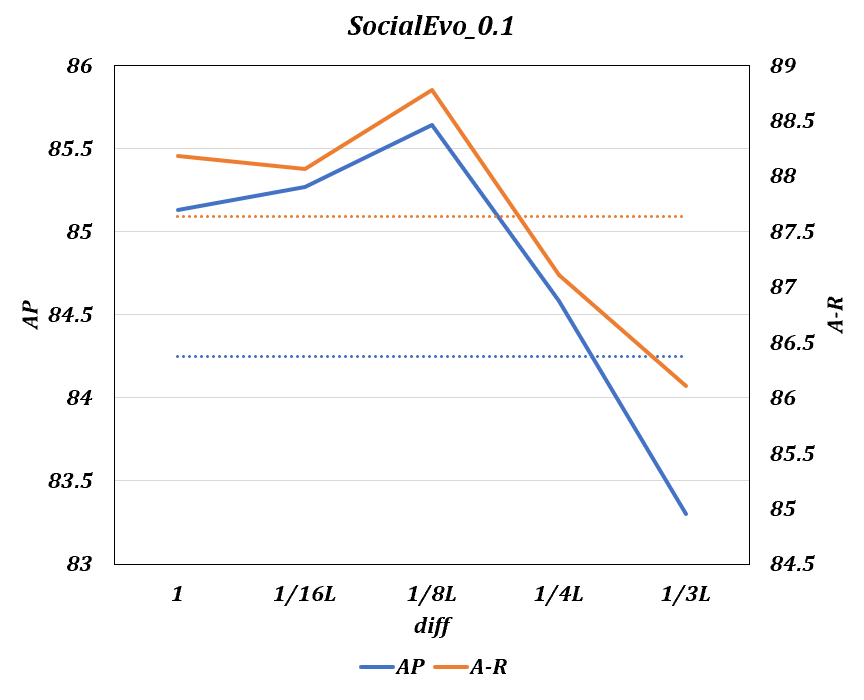}
    \includegraphics[width=0.31\linewidth]{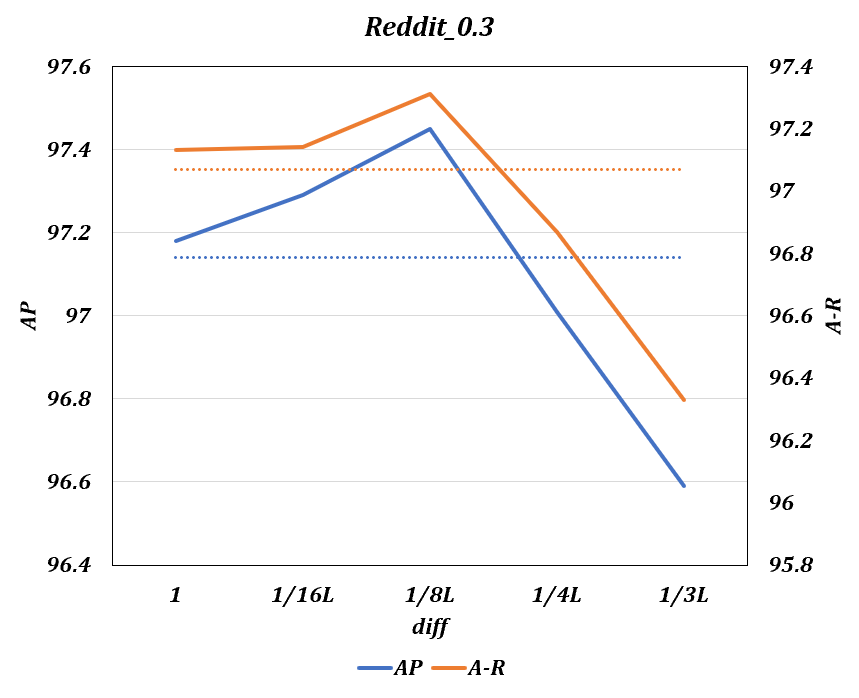}
    \includegraphics[width=0.31\linewidth]{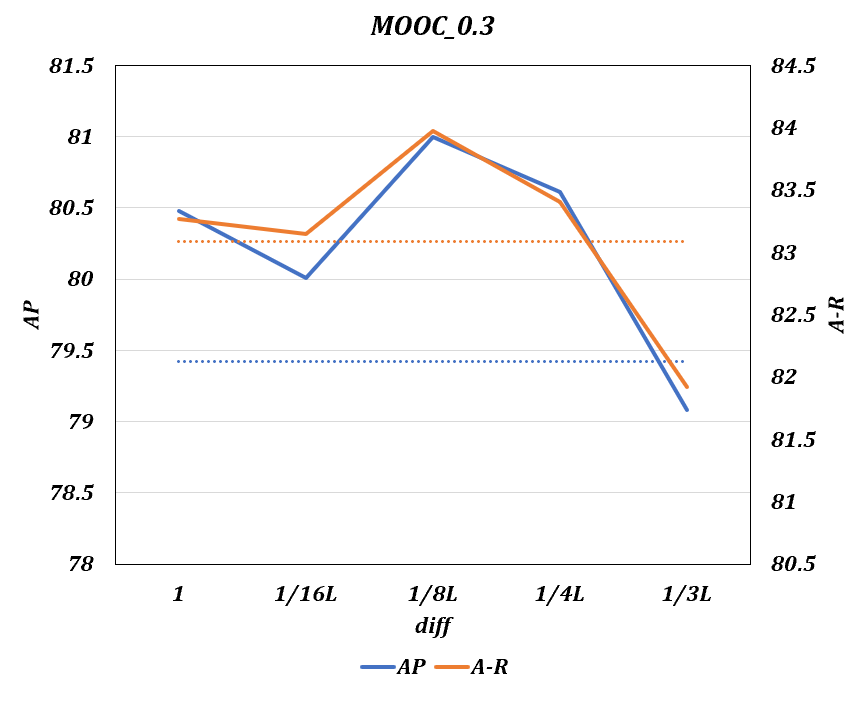}
    \includegraphics[width=0.31\linewidth]{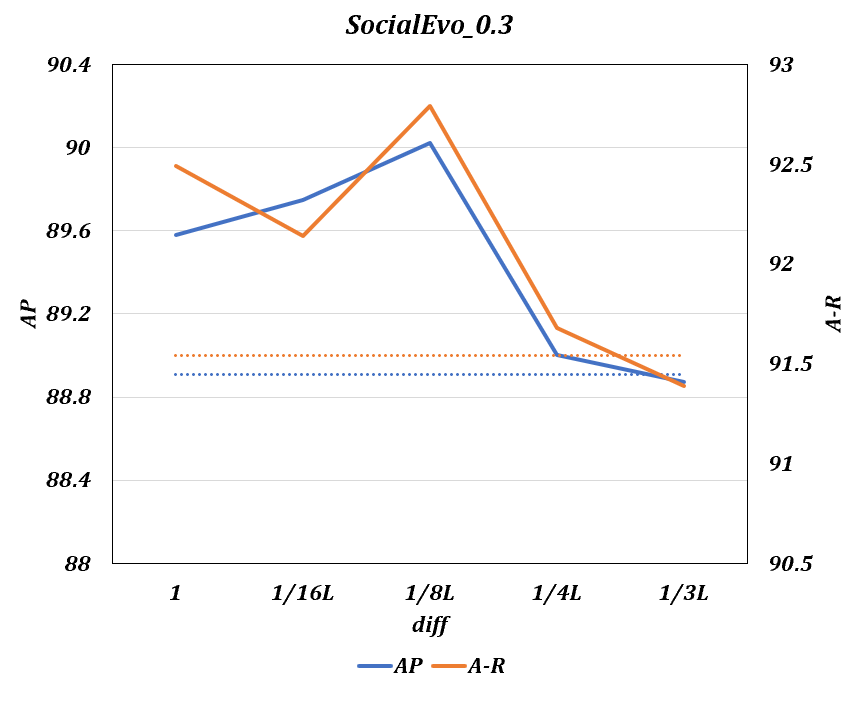}

\caption{Performance comparison of different diffused length $diff$ in DyGFormer+Conda on Reddit, MOOC, SocialEvo with different train set ratios. The blue and orange dashed lines respectively represent the baseline's AP and A-R values.}\label{fig:diff}

\end{figure*}

\begin{figure*}[t]
\centering
    \includegraphics[width=0.31\linewidth]{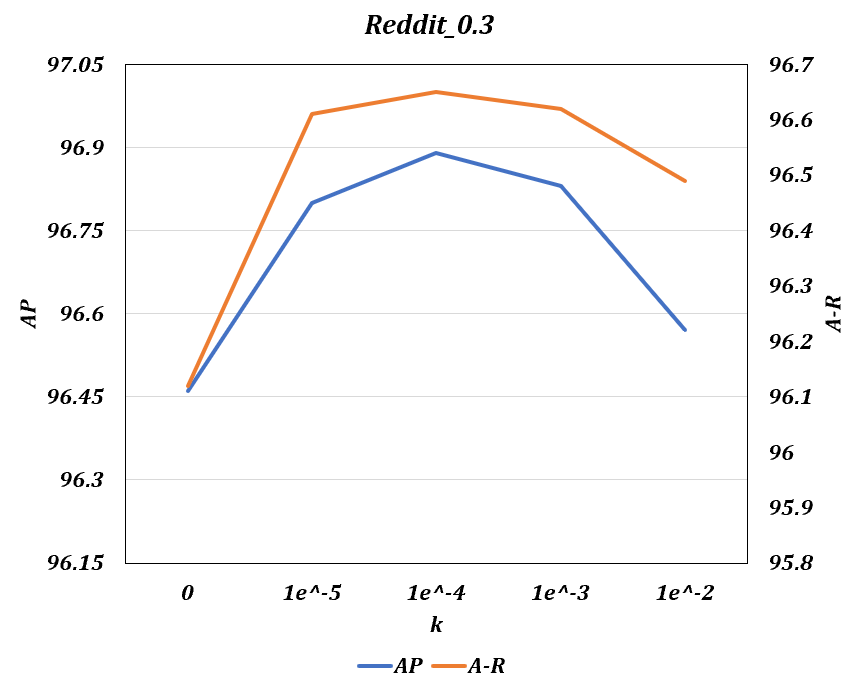}
    \includegraphics[width=0.31\linewidth]{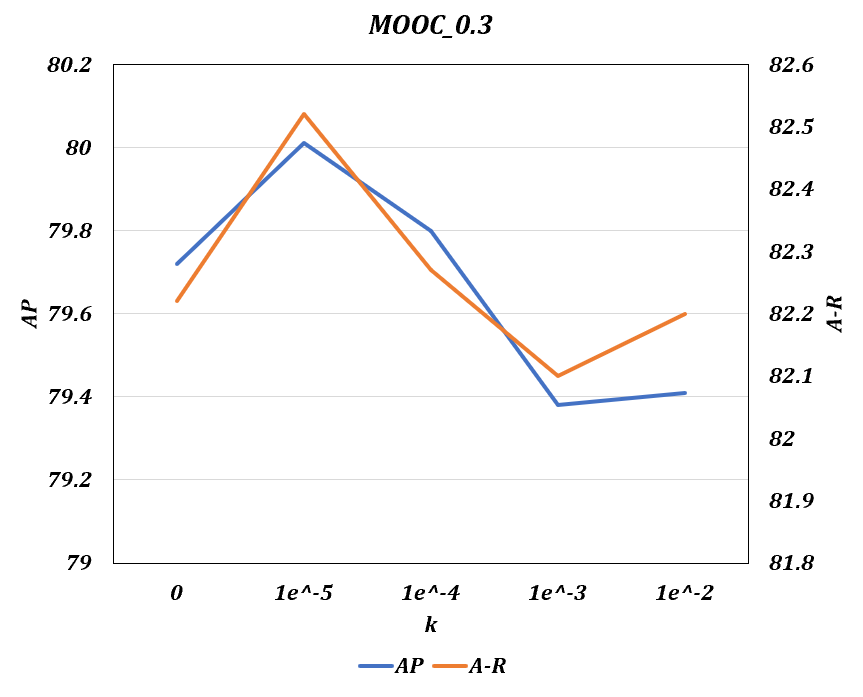}
    \includegraphics[width=0.31\linewidth]{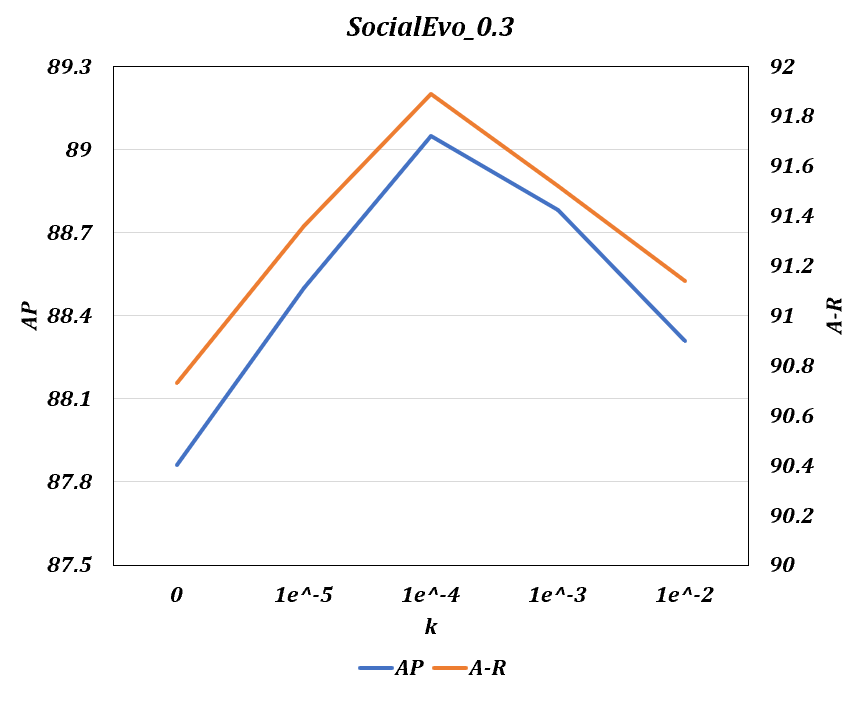}
  
\caption{Performance comparison of different noise scale $k$ in GraphMixer+Conda on Reddit\_0.3, MOOC\_0.3, SocialEvo\_0.3. $k=0$ equivalents to the baseline.}\label{fig:k}

\end{figure*}

\subsection{Sensitivity of Hyper-Parameters}
In this section, we conduct experiments to investigate the Sensitivity of two important hyper-parameters in our proposed method: diffused sequence length $diff$ and noise scale $k$. 

We conducted experiments with DyGFormer on Reddit, MOOC, and SocialEvo datasets, as DyGFormer tends to yield better results from longer historical neighbor sequences on these datasets, which can effectively show the effect on the model performance of using varying diffused lengths across the historical neighbor sequence. We also provide the optimal configurations of the number $L$ of sampled neighbors and diffused sequence length $diff$ of different baselines on different datasets in Appendix~\ref{sec:config}. Specifically, we first set $L$, the number of sampled historical neighbors by DyGFormer on each dataset, to the optimal settings which can be found in the Appendix~\ref{sec:config}. Note that in practice, if the target node's historical neighbors are fewer than $L$, we use zero-padding to fill the gap. Then, we vary $diff$ in the set ${1, \frac{L}{16},\frac{L}{8},\frac{L}{4},\frac{L}{3}}$, with results as shown in the Figure~\ref{fig:diff}. From the Figure, we can observe that the model performance is best when $diff=\frac{L}{8}$. However, as $diff$ increases, such as when $diff=\frac{L}{3}$, the model performance significantly decreases, even falling below the baseline. This phenomenon is particularly pronounced when the training set ratio of the dataset is 0.1. The underlying reason is likely due to the fact that there are few actual neighbors in the sampled historical neighbor sequence of the node, with the latter part of the sequence being filled by zero-padding. Therefore, the majority of the conditional inputs of the reverse process are meaningless, resulting in the generated augmented embeddings being too distant from the current node state representation, leading to a decline in model performance. When $diff$ is less than $\frac{L}{8}$, the model performance is slightly worse than at $\frac{L}{8}$ but still outperforms the baseline. This phenomenon indicates that Conda can consistently generate positive augmentation when it controls the length of diffused sequence embedding to be relatively short.

For the noise scale, we conduct experiments on GraphMixer+Conda with Reddit\_0.3, MOOC\_0.3, and SocialEvo\_0.3 . As illustrated in Figure~\ref{fig:k}, we can observe that, as the noise scale increases, the performance first rises compared to training without noise ($k=0$), verifying the effectiveness of denoising training. However, enlarging noise scales degrades the performance due to corrupting the pattern of interaction sequence. Hence, we also should carefully choose a relatively small noise scale (e.g. $1e^{-4}$).

\section{Related Work}\label{sec:related}

\subsection{Graph Data Augmentation}
There is a growing interest among researchers in graph data augmentation (GDA) methods since they offer an attractive solution in denoising and generally augmenting graph data. GDA methods can be categorized into structure-oriented and feature-oriented methods by the data modality that they aim to manipulate. Structure-oriented GDA methods often modify the graph structure via adding or removing edges and nodes.~\citet{Gaug} and~\citet{GDC} modify the graph structure and used the modified graph for training/inference,~\citet{rong2020dropedge,dropnode} randomly drop edges/nodes from the observed training graph.~\citet{Graphcrop} utilizes a node-centric strategy to crop a subgraph from the original graph while maintaining its connectivity. However, these GDA methods are usually used in static graphs or DTDG and can not be directly applied to CTDG due to the lack of consideration of time, Although~\citet{wang2021adaptive} introduces MeTA for CTDG model, which augments CTDG combining three structure-oriented GDA methods including perturbing time, removing edges, and adding edges with perturbed time. However, it is limited to apply on CTDG models with memory modules~\cite{TGN,DyRep} because it needs to incorporate a multi-level memory module to process augmented graphs of varying magnitudes at different levels. Furthermore, It is widely noticed that the effectiveness structure-oriented GDA methods requires a great of specific domain knowledge, necessitating the selection of diverse augmentation strategy combinations tailored to different graph datasets.


Feature-oriented methods directly modify or create raw features.~\citet{graphmae} uses Attribute Masking that randomly mask node features,~\citet{kong2020flag} augments node features with gradient-based adversarial perturbations. It’s worth noting that structure-oriented and feature-oriented augmentation are also sometimes combined in some GDA methods. For example,~\citet{you2020graph} summarizes four types of graph augmentations to learn the invariant representation across different augmented view.~\citet{Nodeaug} changes both the node feature and the graph structure for different nodes individually and separately, to coordinate DA for different nodes. However, most of these methods require original features for nodes or edges. Meanwhile, Most CTDG datasets are attribute-free graphs. Additionally, there are only a few structure-oriented and feature-oriented GDA methods that offer rigorous proofs or theoretical bounds (e.g., Evidence Lower Bound). Most rely predominantly on empirical intuition or constraints from contrastive learning to achieve positive augmentation.

\subsection{Generative Models}

Generative models~\cite{goodfellow2014generative, kingma2013auto} are powerful tools for learning data distribution. Recently, researchers have proposed several interesting generative models for graph data generation. Variational graph auto-encoder (VGAE)~\citep{kipf2016variational} exploits the latent variables to learn interpretable representations for undirected graphs. \citet{salha2019keep} make use of a simple linear model to replace the GCN encoder in VGAE and reduce the complexity of encoding schemes. \citet{xu2019generative} propose a generative GCN model to learn node representations for growing graphs. ConDgen~\citep{yang2019conditional} exploits the GCN encoder to handle the invariant permutation for conditional structure generation. Besides, diffusion-based generative models are shown to be powerful in generating high-quality graphs~\cite{niu,haefeli2022diffusion,chen2023nvdiff,digress}. DiGress~\cite{digress}, one of the most advanced graph generative models, employs a discrete diffusion process to progressively add discrete noise to graphs by either adding or removing edges and altering node categories. However, these diffusion models rely on continuous Gaussian noise and do not align well with graph structure. In addition, they are limited to generating small graphs and can not scale up to large graphs. Contrary to these approaches mainly focusing on structure generation, ~\cite{Local_aug} pretrains a VAE for node feature generation, which can serve as a DA method for the downstream backbone models. However, VAE often uses over-simplified prior and decoder, which suffers from the trade-off between tractability and representation ability.

\section{Conclusion}
In this paper, we propose Conda, a novel GDA method designed to integrate seamlessly into any CTDG model.
Conda utilizes a latent conditional diffusion model to enhance the embeddings of nodes' historical neighbor sequences during the training phase of CTDG models. 
Unlike structure-oriented and feature-oriented GDA methods, Conda operates within the latent space rather than the input space, thereby enabling more subtle modeling of transition in dynamic graphs. 
More importantly, Conda employs a conditional diffusion model to generate high-quality historical neighbor embeddings with solid theoretical foundations. 
Extensive experiments conducted on various baseline models using real-world datasets demonstrate the efficacy of our method Conda.
In the future, we aim to extend our method to CTDG with edge deletions.

\section{Acknowledgments}
The research presented in this paper is supported in part by the National Natural Science Foundation of China (Grant No. 62372362) and the National Natural Science Foundation of China (Grant No. 62366021).

\bibliographystyle{ACM-Reference-Format}
\bibliography{Conda}

\begin{appendices}

\section{Appendix}
\begin{table*}[t]
    \renewcommand\arraystretch{1.2}
    \setlength{\tabcolsep}{3.0mm}{\small
    \begin{tabular}{cccccccc}
    \hline
    \multicolumn{1}{c}{Dataset}  & Nodes  & Edges  & Unique Edges & Node/Link Feature &Time Granularity &Duration&density \\ \hline
    WIKI                   & 9227  & 157474 &18257 & 0/172 &Unix timestamp &1 month &4.30E-03 \\
    REDDIT                 & 10984 & 672447 &78516 & 0/172 &Unix timestamp &1 month &8.51E-03 \\
    MOOC                    & 7144  & 411749 &178443 & 0/4 &Unix timestamp &17 month &1.26E-02 \\
    LastFM  &1980 &1293103 &154993 &0/0 &Unix timestamp &1 month &5.57E-01\\
    Social Evo.  &74 &2099519 &4486 &0/2 &Unix timestamp & 8 months &5.36E+02\\
    UCI  &1899 &59835 &20296&0/0 &Unix timestamp & 196 days &3.66E-02\\\hline
    \end{tabular}}
    \caption{Dataset statistics}\label{tab:statTable}
\end{table*}
\subsection{Detail Descriptions of Datasets}\label{sec:dataset}

\begin{itemize}[leftmargin=*]
    \item \textbf{Wiki}: is a bipartite interaction graph that contains the edits on Wikipedia pages over
    a month. Nodes represent users and wiki pages, and links denote the editing behaviors with
    timestamps. Each link is associated with a 172-dimensional Linguistic Inquiry and Word
    Count (LIWC) feature. 
    
    \item \textbf{Reddit}: consists of one month of posts made by users on subreddits. Users
    and subreddits are the nodes, and links are the timestamped posting requests. Each link has a 172-dimensional LIWC feature. 
    
    \item \textbf{MOOC}: is a bipartite interaction network of online sources, where nodes are students and course content units (e.g., videos and problem sets). Each link denotes a student’s access behavior to a specific content unit and is assigned a 4-dimensional feature.
    
    \item \textbf{UCI}: is a Facebook-like, unattributed online communication network among students of the University of California at Irvine, along with timestamps with the temporal granularity of seconds.
     
    \item \textbf{LastFM}: is an interaction network where users and songs are nodes and each edge represents a user-listens-to-song relation. The dataset consists of the relations of 1000 users listening to the 1000 most listened songs over a period of one month. The dataset contains no attributes.
     
    \item \textbf{SocialEvo}: is a mobile phone proximity network that tracks the everyday life of a whole undergraduate dormitory from October 2008 to May 2009. Each edge has 2 features.
\end{itemize}

\subsection{Detail Descriptions of baselines}\label{sec:baselines}
    \begin{itemize}[leftmargin=*]
    \item \textbf{JODIE} is a RNN-based method. Denote $h_u(t)$ as the embedding of node $u$ at timestamp $t$, $e_{u,v}^t$ as the link feature between $u$, $v$ at timestamp $t$, and $m_u$ as the timestamp that node $u$ latest interact with other nodes. 
    When an interaction between node $u$, $v$ happens at timestamp $t$, JODIE updates the node embedding using RNN by $h_u(t) = RNN(h_u(m_u),h_v(m_v),e_{u,v}^t,t - m_u)$. Then, the embedding of node $u$ at timestamp $t_0$ is computed as $h_u(t_0) = (1 + (t_0 - m_u)w) \cdot h_u(m_u)$.
    
    \item \textbf{TGAT} is a self-attention-based method that could capture spatial and temporal information simultaneously. TGAT first concatenates the raw feature $x_u$ with a trainable time encoding $z(t)$, i.e., $x_u(t) = [x_u || z(t)]$ and $z(t) = cos(tw+b)$. Then, self-attention is applied to produce node representation $h_u(t_0) = SAM (x_u(t_0), {x_v(m_v) | v \in N_{t_0}(u)})$, where $N_{t_0}(u)$ denotes the neighbors of node $u$ at time $t_0$ and $m_u$ denotes the timestamp of the latest interaction of node $u$. Finally, the prediction on any node pair at time $t_0$ is computed by $MLP([h_u(t_0) || h_v(t_0)])$. 
    
    \item \textbf{TGN} is a mixture of RNN- and self-attention based method. TGN utilizes a memory module to store and update the (memory) state $s_u(t)$ of node $u$. The state of node $u$ is expected to represent $u$’s history in a compressed format. Given the memory updater as mem, when an link $e_{uv}(t)$ connecting node $u$ is observed, node $u$’s state is updated as $s_u(t) = mem(s_u(t^-),s_v(t^-)||e_{uv}(t))$. where $s_u(t^-)$ is the memory state of node $i$ just before time $t$. $||$ is the concatenation operator, and node $v$ is $u$’s neighbor connected by $e_{u,v}(t)$. The implementation of $mem$ is a recurrent neural network (RNN), and node $u$’s embedding is computed by aggregating information from its L-hop temporal neighborhood using self-attention.

    \item \textbf{DyRep} is an RNN-based method that updates node states upon each interaction. It also includes a temporal-attentive aggregation module to consider the temporally evolving structural information in dynamic graphs.

     \item \textbf{TCL} is a contrastive learning-based method. It first generates each node’s interaction sequence by performing a breadth-first search algorithm on the temporal dependency interaction sub-graph. Then, it presents a graph transformer that considers both graph topology and temporal information to learn node representations. It also incorporates a cross-attention operation for modeling the inter-dependencies of two interaction nodes.
     
      \item \textbf{GraphMixer} is a simple MLP-based architecture. It uses a fixed time encoding function that performs rather than the trainable version and incorporates it into a link encoder based on MLP-Mixer to learn from temporal links. A node encoder with neighbor mean-pooing is employed to summarize node features.
    
    \item \textbf{DyGFormer} is a self attetion-based method. Specifically, for node $n_i$, DyGFormer just retrieves the features of involved neighbors and links based on the given features to represent their encodings. DyGFormer is equipped with a neighbor co-occurrence encoding scheme, which encodes the appearing frequencies of each neighbor in the sequences of the source node and destination node, and can explicitly explore the correlations between two nodes. Instead of learning at the interaction level, DyGFormer splits each source/destination node’s sequence into multiple patches and then feeds them to the transformer. 
    
    \end{itemize}
    
\subsection{Descriptions of GDA Methods in experiments}\label{sec:GDAmethods}
\begin{itemize}[leftmargin=*]
    \item \textbf{DropEdge} which randomly drops edges according to the drop possibility $p$ at data pre-process phase. This slight modification of the original graph results in the GNN observing a different graph at each epoch.  
    \item \textbf{DropNode} Similar to DropEdge, DropNode drops nodes according to the drop possibility $p$ at the data pre-process phase.
    \item \textbf{MeTA} a dynamic graph data augmentation module that stacks a few levels of memory modules to augment dynamic graphs of different magnitudes on separate levels with three data augmentation strategies, including perturbing time, removing edges, and adding edges with perturbed time. 
\end{itemize}

\subsection{Detail configurations}\label{sec:config}

\begin{table}[h]
    \renewcommand\arraystretch{2.2}
    \setlength{\tabcolsep}{0.8mm}{\footnotesize
    \begin{tabular}{c|ccccccc}
    \hline
      Dataset& Wiki\_0.3& Reddit\_0.3& MOOC\_0.3 & UCI\_0.3& LastFM\_0.3&SocialEvo\_0.3  \\ \hline
      JODIE & 10/$\frac{L}{4}$ &10/$\frac{L}{4}$  & 10/$\frac{L}{4}$ & 10/1 &10/$\frac{L}{3}$  &10/$\frac{L}{3} $   \\
      DyRep&10 /$\frac{L}{4}$& 10/$\frac{L}{4}$ & 10/$\frac{L}{4}$ & 10/1 & 10/$\frac{L}{3}$ &10/$\frac{L}{3}$   \\
      TGAT & 20 /$\frac{L}{4}$ &20/$\frac{L}{4}$ &20 /$\frac{L}{8}$  & 20/$\frac{L}{8}$ & 20/$\frac{L}{8}$ &20/$\frac{L}{4}$   \\
      TGN & 10 /$\frac{L}{4}$ &10 /$\frac{L}{3}$& 10/$\frac{L}{4}$ & 10/$\frac{L}{4}$ & 10/$\frac{L}{4}$ &10/$\frac{L}{3}$    \\
      TCL & 20/$\frac{L}{8}$ &20/$\frac{L}{8}$ & 20/$\frac{L}{8}$ & 20/$\frac{L}{8}$ &20/$\frac{L}{8}$  &20/$\frac{L}{4}$    \\
      GraphMixer & 32/$\frac{L}{8}$& 10/$\frac{L}{3}$ &20/$\frac{L}{4}$  & 20/$\frac{L}{8}$ & 10/1 &20/$\frac{L}{4}$    \\ 
      DyGFormer & 32/$\frac{L}{4}$& 64/$\frac{L}{8}$& 128/$\frac{L}{8}$ & 32/$\frac{L}{8}$ & 256/$\frac{L}{8}$ & 32/$\frac{L}{4}$   \\ \hline
    \end{tabular}}
    \caption{The optimal configurations of the number $L$ of sampled neighbors and length $diff$ of diffused sequence}\label{tab:num_sample_0.3}
\end{table}
    
\begin{table}[h]
\renewcommand\arraystretch{2.2}
\setlength{\tabcolsep}{0.8mm}{\footnotesize
\begin{tabular}{c|ccccccc}
\hline
  Dataset& Wiki\_0.1& Reddit\_0.1& MOOC\_0.1 & UCI\_0.1& LastFM\_0.1&SocialEvo\_0.1  \\ \hline
  JODIE & 10/$\frac{L}{4}$&10/$\frac{L}{4}$ & 10/$\frac{L}{4}$ & 10/1 &10/1  &10/$\frac{L}{4}$    \\
  DyRep&10/1 & 10/1 & 10/1 & 10/1 & 10/1&10/1   \\
  TGAT & 20/$\frac{L}{8}$&20/$\frac{L}{8}$ &20/$\frac{L}{8}$  & 20/$\frac{L}{8}$ & 20/$\frac{L}{8}$ &20/$\frac{L}{8}$   \\
  TGN & 10/1 &10/$\frac{L}{4}$& 10/$\frac{L}{4}$ & 10/1 & 10/1 &10/$\frac{L}{4}$    \\
  TCL & 20/$\frac{L}{8}$&20/$\frac{L}{8}$ & 20/$\frac{L}{8}$ & 20/1 &20/$\frac{L}{8}$  &20/$\frac{L}{4}$    \\
  GraphMixer & 20/$\frac{L}{8}$& 10/$\frac{L}{8}$ &10/$\frac{L}{4}$  & 10/$\frac{L}{8}$ & 10/$\frac{L}{8}$ &20/$\frac{L}{8}$    \\ 
  DyGFormer & 32/$\frac{L}{8}$& 20/$\frac{L}{8}$& 32/$\frac{L}{8}$ & 32/1 & 64/$\frac{L}{16}$ & 32/$\frac{L}{8}$   \\ \hline
\end{tabular}}
\caption{The optimal configurations of the number $L$ of sampled neighbors and length $diff$ of diffused sequence}\label{tab:num_sample_0.3}
\end{table}

\end{appendices}

\end{document}